\pdfoutput=1
\documentclass[11pt]{article}

\usepackage[]{acl}

\usepackage{times}
\usepackage{latexsym}

\usepackage[T1]{fontenc}
\usepackage[utf8]{inputenc}

\usepackage{microtype}

\usepackage{graphicx}
\usepackage{soul}
\usepackage{colortbl}
\usepackage{multicol}
\usepackage{booktabs}
\usepackage{amsmath}
\usepackage{xspace}

\usepackage[utf8]{inputenc} % allow utf-8 input
\usepackage[T1]{fontenc}    % use 8-bit T1 fonts
\usepackage{arydshln}

\definecolor{lred}{rgb}{1.000, 0.773, 0.882}
\definecolor{lbue}{rgb}{0.757, 0.882, 0.992}
\definecolor{lgreen}{rgb}{0.878, 1.000 , 0.784}

\newcommand{\cred}[1]{\protect\sethlcolor{lred}\protect\hl{#1}}
\newcommand{\cblue}[1]{\protect\sethlcolor{lbue}\protect\hl{#1}}
\newcommand{\cgreen}[1]{\protect\sethlcolor{lgreen}\protect\hl{#1}}

\newcommand{\dialogsum}{\emph{DialogSum}\xspace}

\newcolumntype{C}[1]{>{\centering\arraybackslash}m{#1}}
\newcolumntype{L}[1]{>{\raggedright\arraybackslash}m{#1}}

\title{\emph{DialogSum Challenge}: Results of Dialogue Summarization Shared Task}

\author{
Yulong Chen \thanks{\ \ Equal Contribution.} $^{\ 1}$, Naihao Deng $^{*\ 2}$, Yang Liu, Yue Zhang $^{1, 3}$\\
$^1$ School of Engineering, Westlake University\\
$^2$ School of Computer Science and Engineering, University of Michigan\\
$^3$ Institute of Advanced Technology, Westlake Institute for Advanced Study\\
\emph{\href{mailto:yulongchen1010@gmail.com}{yulongchen1010@gmail.com}}
\quad\emph{\href{mailto:dnaihao@umich.edu}{dnaihao@umich.edu}}\\
\emph{\href{mailto:inf.yangl@outlook.com}{inf.yangl@outlook.com}}\quad\emph{\href{mailto:yue.zhang@wias.org.cn}{yue.zhang@wias.org.cn}}
}

\begin{document}
\maketitle
\begin{abstract}
We report the results of \emph{DialogSum Challenge}, the shared task on summarizing real-life scenario dialogues at INLG 2022.
Four teams participate in this shared task and three submit their system reports, exploring different methods to improve the performance of dialogue summarization.
Although there is a great improvement over the baseline models regarding automatic evaluation metrics, such as \textsc{Rouge} scores, we find that there is a salient gap between model generated outputs and human annotated summaries by human evaluation from multiple aspects.
These findings demonstrate the difficulty of dialogue summarization and suggest that more fine-grained evaluatuion metrics are in need.
\end{abstract}

\begin{figure*}
    \centering
    \includegraphics[width=0.95\linewidth]{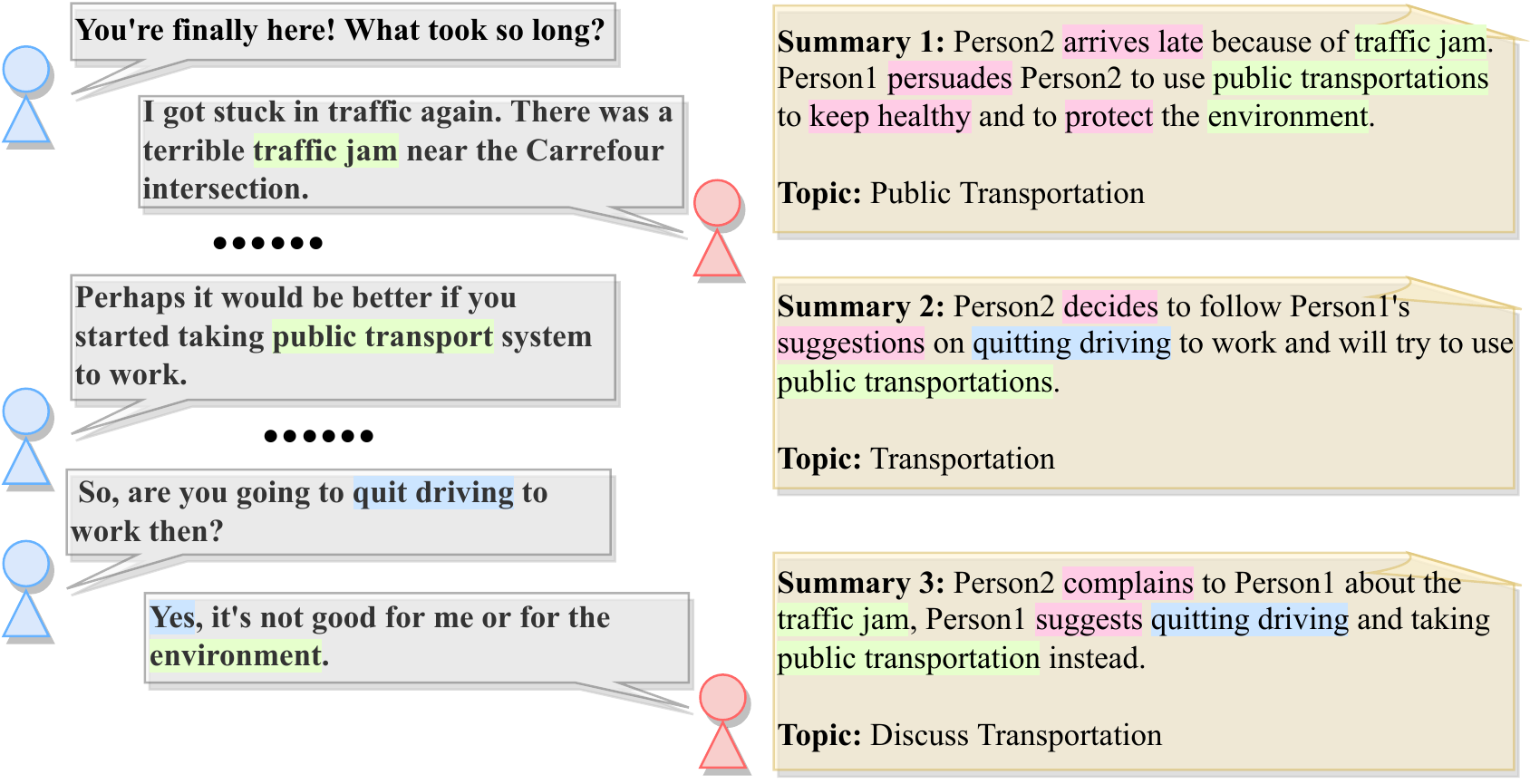}
    \caption{An example in the public test set of the \dialogsum dataset.
    Tokens highlighted in red represent \cred{tokens that only appear in the summary but not in the dialogue text}, requiring the model to summarize with a high level semantic understanding. 
    Tokens hightlighted in blue represent \cblue{the information that spans across turns}. Tokens highlighted in green show \cgreen{the corresponding information in the dialogue text and the summary}. 
    Note that such information scatters in various places in the dialogue.}
    \label{fig: dialogsum-dataset-example}
\end{figure*}

\section{Introduction}

With the power of Pretrained Language Models (PLMs), research on text summarization has made great progresses~\cite{liu2019text}. 
However, previous research focuses on monologue summarization, such as news articles~\cite{paulus2017deep, gehrmann2018bottom, liu2019text, liu2020noisy}, patents~\cite{subramanian2019extractive} and academic papers~\cite{koncel2019text}.
However, as an important communicative channel~\cite{bender2020climbing}, dialogues receive less attention from the community. 

To this end, we propose \emph{DialogSum Challenge} to encourage researchers to investigate different solutuons for real-life summarization~\cite{chen-etal-2021-dialogsum-challenge}. 
Different from previous dialogue summarization tasks~\cite{carletta2005ami, gliwa2019samsum}, \emph{DialogSum Challenge} focuses on diverse real-life scenarios such as schooling, work, medication, shopping, leisure, travel with large scale data.

The challenges of \emph{DialogSum} can be stated from three perspectives.
First, \emph{DialogSum} include a variety of topics, requiring models to process text  with different real-life scenarios.
Second, compared with well-structured monologues, dialogues have unique discourse structures and language styles~\cite{grosz1995centering}.
The structures and use of languages differ from monologues, for instance, the key information spans over the context~\cite{grosz-etal-1995-centering}, which makes dialogues more difficult to encode. 
Third, compared with monologue summarization, dialogue summaries are written from a different perspective, usually including speakers' intents and actions~\cite{chen2021dialogsum}.
Therefore, dialogue summarization is abstractive in nature and requires a high level understanding beyond text semantics~\cite{chopra2016abstractive, khandelwal2019sample}.
Figure~\ref{fig: dialogsum-dataset-example} shows an example in \emph{DialogSum}.
Apart from the research challenges, \emph{DialogSum Challenge} incentivizes summarization systems that can help end users. 
Summarizing daily spoken dialogues can help archive the important information in business and personal communication. 
This presumably lifts the burden of manually taking notes, liberating human beings from the tedious work.

Responding to our calls, four teams participate in the shared task, and three of them submit their system reports.
The submitted systems typically employ PLMs, such as BART~\cite{lewis2019bart} and \textsc{Pegasus}~\cite{zhang2020pegasus}.
In addition, they explore diverse methods to improve the performance, including integrating additional features, modifying the decoding process for better summary generation, multi-task tuning the model with auxiliary tasks and using data from other sources.

To evaluate the performance, we construct a hidden test set that contains 100 manually annotated samples, and evaluate models on hidden and public test sets using both automatic and manual evaluations.
For automatic evaluation, we use \textsc{Rouge} scores and \textsc{BERTScore}.
For manual evaluation, we follow \emph{DialogSum Challenge}~\cite{chen-etal-2021-dialogsum-challenge} and evaluate model outputs from multiple aspects.
Results show that tuning models on CNN/Daily News corpus~\cite{hermann2015teaching} or AMI dataset~\cite{carletta2005ami}, and incorporating topics in summary generation process can improve the model performance. However, there are still rooms for models to improve the metric scores, as well as the quality of the generated summaries in terms of identifying the interlocutors' intents, capturing the discourse relation, etc. Besides, we observe the mismatch between \textsc{BERTScore} and human scores, which aligns with the findings by~\citet{hanna2021fine}.

Full details of the shared task description and logistics, as well as the dataset can be found at \url{https://cylnlp.github.io/dialogsum-challenge/}.

\section{Task}
Given a piece of dialogue text as input, the task is to ask a model to generate a summary that conveys the key information of the dialogue.
The output summary should be concise, coherent, consistent and be written from a listener's perspective.

\section{Data}

\begin{table*}[t]
    \centering
    \begin{tabular}{lcccccccc}
        \toprule
        & \multicolumn{4} {c} {Public Test Set} & \multicolumn{4} {c} {Hidden Test Set}\\
         Model & R1 & R2 & RL & \textsc{BERTScore} & R1 & R2 & RL & \textsc{BERTScore}\\
         \midrule
         Human & \textbf{53.35} & \textbf{26.72} & \textbf{50.84} & 92.63 & - & - & - & - \\
         \hdashline
         GoodBai & \textbf{47.61} & \textbf{21.66} & 45.48 & \textbf{92.72} & 49.66 &	\textbf{26.03} &	\textbf{48.55} &	91.69 \\
         UoT & 47.29 &	21.65 &	\textbf{45.92} &	92.26 &	\textbf{49.75} &	25.15 &	46.50 &	\textbf{91.76}\\
         IITP-CUNI & 47.26 &	21.18 &	45.17 &	92.70 &	45.89 &	21.88 &	43.16 &	91.13\\
         \bottomrule
         
    \end{tabular}
     \caption{Scores by automatic metrics for each submission and human results. We embolden the top scores among models, as well as the human score if it is the highest among all the scores.}
    \label{tab: submission-result-automatic-metrics}
\end{table*}

\begin{table*}[t]
    \centering
    \begin{tabular}{lcccccccc}
        \toprule
        & \multicolumn{4} {c} {Public Test Set} & \multicolumn{4} {c} {Hidden Test Set}\\
         Model & R1 & R2 & RL & \textsc{BERTScore} & R1 & R2 & RL & \textsc{BERTScore}\\
                 \midrule
         TCS\_WITM & 47.02 &	21.20 &	44.90 &	90.13 &	50.32 &	25.59 &	47.40 &	91.81\\
         \bottomrule
         
    \end{tabular}
     \caption{Scores by automatic metrics for the submission from TCS\_WITM. The model submitted by TCS\_WITM predicts 3 summaries based on the 3 topics in the public test set. We take the highest score among the 3 summaries to calculate the scores.}
    \label{tab: submission-result-automatic-metrics-tcs}
\end{table*}

\paragraph{Data Sources}
We use the train, dev and public test data from \dialogsum. Additionally, we collect 100 summaries as hidden test set from the same website where \dialogsum crawls the data~\footnote{\url{http://tingroom.com}}. We follow the exact same procedure as the annotation for the original \dialogsum dataset~\cite{chen2021dialogsum}. We remove the non-English characters, correct typos and grammatical errors, and filter out duplicated dialogues based on text similarity~\footnote{We compute the \textsc{Rouge} scores between two dialogues and filter out dialogues that have more than $80\%$ \textsc{Rouge-1} scores.}.
The annotators are instructed to write the summaries for each dialogue by: (1) conveying the salient information in the dialogue and; (2) keeping the summary short and; (3) writing from the observer perspective and in formal language.
Additionally, we ask annotators to keep tense consistency, preserve important discourse relations, explicitly describe emotion and speaker's intents. 
Also, annotators are instructed to provide a short topic for each dialogue.
Table~\ref{tab: dataset-splits} shows the statistics for the data in \emph{DialogSum Challenge}.

\begin{table}[t]
\centering
\begin{tabular}{cccc|c}
\toprule
Train  & Dev & Test$_\text{public}$ & Test$_\text{hidden}$ & Total\\
\midrule
12,460 & 500 & 500                  & 100               & 13,560  \\
\bottomrule
\end{tabular}
\caption{Number of dialogues in each split for \emph{DialogSum Challenge}.}
\label{tab: dataset-splits}
\end{table}

\section{Evaluation Set-Ups}
\paragraph{Automatic Evaluation}
We adopt two metrics, \textsc{Rouge}~\cite{lin2004rouge} and \textsc{BERTScore}~\cite{zhang2019bertscore} for automatic evaluation. We use RoBERTa~\cite{liu2019roberta} large as the backbone to calculate \textsc{BERTScore}. 

\paragraph{Manual Evaluation}
Furthermore, we conduct manual evaluation from various aspects, including standard summarization metrics~\citet{kryscinski2019neural, kryscinski2019evaluating}, coreference information, intent identification, discourse relation following~\citet{chen2021dialogsum}, as well as objectiveness (whether the summary is insusceptible to subjectivity such as subjective assumptions in the dialogues). Besides, annotators give an overview score for the predicted summary.

\section{Submissions}

\subsection{IITP-CUNI}

The model submitted by Indian Institute of Technology Patna and Charles University employs a multi-task learning set-up to improve model performance.
In their experiments, they explore several auxiliary tasks including extractive summarization to classify whether a given sentence belongs to the summary or not, novelty detection~\cite{ghosal2022novelty} of whether the given summaries correspond to the same dialogue, as well as a masked language modeling~\cite{devlin2018bert} task to recover summaries.
They find that the BART~\cite{lewis2019bart} large model tuned with the auxiliary task of the extractive summarization task with  data from AMI~\cite{carletta2005ami} corpus performs the best.

\subsection{UoT}

The participants from the University of T\"ubingen use the pre-trained BART model which is further tuned on CNN/Daily News corpus~\cite{hermann2015teaching}. 
Besides, they penalize generating longer summaries in the decoding part of the model, and post-process the summaries to resolve generation errors (e.g. replacing speakers' names who do not appear in the dialogue with \#Person\_1\# or \#Person\_2\#, and fixing duplicated labels such as \#Person\_1\#Person\_1\# to \#Person\_1\#). 

They also explore methods such as intermediate task transfer learning with training on commonsense reasoning task or other summarization task first and then tune the model on \dialogsum, transforming dialogue structures to news with structures similar to what BART model is trained on,  as well as data augmentation by using data from SAMSum~\cite{gliwa2019samsum} dataset. However, they do not find any improvement using these techniques.

\subsection{TCS\_WITM}

The model from TCS research adopts the pre-trained PEGASUS~\cite{zhang2020pegasus} large model which is further fine-tuned on CNN/Daily News. They also incorporate the topics provided in \dialogsum dataset and feed the topic together with the dialogue text to their model. The extra information from the topics boost up the model performance compared to the baseline performance of simply feeding dialogues to the model.

\begin{figure}[t]
    \centering
    \begin{tabular}{L{7.2cm}}
    \toprule
        {\bf Dialogue:} \\
            \#Person1\#: Excuse me, could you tell me how to get to the school clinic? I've \cred{lost my way.} \\
            \#Person2\#: Yes. \cgreen{Go straight ahead till you come to the traffic lights, turn left there and it's the first turning on the right.} \\
            \#Person1\#: Straight ahead to the traffic lights, left and then right. \\
            \#Person2\#: That's it. It'll take you about five minutes. \\
            \#Person1\#: Thank you very much. \\
        \midrule
        {\bf Gold: }\\
        {\it Summary:} \cred{\#Person1\# is lost} on the way to the school clinic. \cgreen{\#Person2\# shows \#Person1\# the correct direction.} \\
        \midrule
        {\bf IIPT-CUNI:} \\
        {\it Prediction: } \cgreen{\#Person2\# shows \#Person1\# the way to the school clinic.}\\
        {\bf UoT: } \\
        {\it Prediction: } \cgreen{\#Person2\# shows \#Person1\# the way to the school clinic.}\\
        {\bf TCS\_WITM: } \\
        {\it Prediction: } \cgreen{\#Person2\# tells \#Person1\# how to get to the school clinic.}\\
        {\bf GoodBai: } \\
        {\it Prediction:} \cgreen{\#Person2\# tells \#Person1\# how to get to the school clinic.}\\ 
    \bottomrule
    \end{tabular}
    \caption{An example where all the predicted summaries miss the the context \cred{\#Person1\# is lost}, while all of the gold summaries contain this context. However, all the predicted summaries successfully capture the event of \cgreen{\#Person2\# shows \#Person1\# the direction}.}
    \label{fig:no-cause-example}
\end{figure}

\section{Results}

Table~\ref{tab: submission-result-automatic-metrics} and Table~\ref{tab: submission-result-automatic-metrics-tcs} show the results of human agreement and submissions from participants in \emph{DialogSum Challenge} by the automatic metrics of \textsc{Rouge}-1 (R1), \textsc{Rouge}-2 (R2), \textsc{Rouge}-L (RL), and \textsc{BERTScore}. 
We do not compare TCS\_WITM with other models because TCS\_WITM uses gold topic information.

In general, scores of the submissions are higher than the baseline models in the original \dialogsum paper, demonstrating the effort from the participants.
Both submissions from UoT and IITP-CUNI tune their models on other dataset then on \dialogsum. 
In particular, UoT tunes their model on CNN/Daily News corpus, while IITP-CUNI tunes their model on AMI dataset.
The reported results show that the model by UoT outperforms the model by IITP-CUNI.
This might be attributed to the different in training size, where AMI~\cite{carletta2005ami} has 137 meetings, while CNN/Daily News corpus~\cite{hermann2015teaching} has 312,000 articles. Thus, the model tuned on CNN/Daily News corpus might have better generalization ability. The model by TCS\_WITM which adopts such a method achieves 50.32 in \textsc{Rouge}-1 score for the hidden test case, showing that generating the summary with the given topic can also help the performance.

However, even the best-performed model underperforms humans by a margin larger than 5.0 in terms of all the \textsc{Rouge} scores. This indicates that \dialogsum is challenging and there is still a large room for future improvement. 

Although existing works on summarization adopt \textsc{BERTScore}~\cite{gabriel2019discourse}, we observe that the \textsc{BERTScore} deviates from the human scores. For instance, though GoodBai achieves the best \textsc{BERTScore} on the public test set, it is TCS\_WITM with a lower \textsc{BERTScore} that achieves the best human scores (Overview score in Table~\ref{tab: submission-result-human-analysis}). Same phenomenon also exists for the hidden test set. This observation aligns with the finding from~\citet{hanna2021fine} that \textsc{BERTScore} performance still deviates from human. Thus, the \textsc{BERTScore} is still not perfect to serve as the ultimate metric for summarization tasks, and our community might come up with a better automatic metric that aligns with human scores.

\begin{table*}[t]
    \small
    \centering
    \begin{tabular}{lcccccc|cccccc}
        \toprule
         & \multicolumn{6}{c|}{Public Test Set} & \multicolumn{6}{c}{Hidden Test Set}\\
         Model & CoRef & Dis & Obj & Intent & Summ & Over & CoRef & Dis & Obj & Intent & Summ & Over \\
         \midrule
         Perfect Score & 1.00 & 1.00 & 1.00 & 1.00 & 5.00 & 5.00 & 1.00 & 1.00 & 1.00 & 1.00 & 5.00 & 5.00\\
         \hdashline
         GoodBai & 0.96 & 0.86 & \textbf{1.00} & 0.72 & 4.12 & 3.96 & \textbf{0.90} & \textbf{0.90} & \textbf{1.00} & 0.70 & \textbf{4.20} & \textbf{4.15} \\
         UoT & \textbf{0.98} & \textbf{0.92} & \textbf{1.00} & \textbf{0.80} & \textbf{4.18} & \textbf{4.08} & 0.75 & 0.75 & \textbf{1.00} & \textbf{0.80} & 4.00 & 3.70\\
         IITP-CUNI & 0.88 & 0.66 & 0.96 & 0.76 & 3.94 & 3.64 & 0.75 & 0.85 & \textbf{1.00} & 0.70 & 3.80 & 3.70\\
         \bottomrule
    \end{tabular}
     \caption{Prediction results by one of the annotators of the \dialogsum dataset. ``CoRef'', ``Dis'', ``Obj'', ``Intent'', ``Summ'', ``Over'' indicates coreference information, discourse relation, objective description, intent identification, standard summarization metrics and overall scores, respectively. We embolden the best scores for each column.}
    \label{tab: submission-result-human-analysis}
\end{table*}

\begin{table*}[t]
    \small
    \centering
    \begin{tabular}{lcccccc|cccccc}
        \toprule
         & \multicolumn{6}{c|}{Public Test Set} & \multicolumn{6}{c}{Hidden Test Set}\\
         Model & CoRef & Dis & Obj & Intent & Summ & Over & CoRef & Dis & Obj & Intent & Summ & Over \\
         \midrule
         TCS\_WITM & 0.88 & 0.90 & 1.00 & 0.82 & 4.20 & 4.10 & 0.90 & 0.80 & 0.84 & 0.70 & 3.95 & 3.80\\
         \bottomrule
    \end{tabular}
     \caption{Prediction results by one of the annotators of the \dialogsum dataset for TCS\_WITM.}
    \label{tab: submission-result-human-analysis-tcs}
\end{table*}
\section{Human Analysis}
\label{sec: human-analysis}

We randomly sample 50 examples for the public test set and 20 examples for the hidden test set to conduct manual analysis.
As discussed in our proposal, we include the metrics of coreference information, discourse relation, objective description, intent identification, standard summarization metrics and overview scores. We annotate -1, 0, 1 for the metrics of coreference information, discourse relation, objective description, intent identification where 1 means all correct, 0 means partially correct and -1 means all incorrect. We annotate from 1 to 5 for the standard summarization metrics and overview scores. The higher, the better.

\paragraph{Coreference Information} Whether the summary aligns interlocutors and their conversation actions or contents.

\paragraph{Discourse Relation} Whether the summary captures important relations between main events, identifying discourse relations and using appropriate phrases to express such relations.

\paragraph{Objective Description} Whether the summary employs objective languages to describe dialogues.

\paragraph{Intent Identification} Whether the summary captures the interlocutors' intents.

\paragraph{Standard Summarization Metrics}~\cite{kryscinski2019neural, kryscinski2019evaluating} Whether the summary is fluent, consistent, relevant and coherent.
However, in practice, we find that summaries generated by PLMs are mostly fluent, sometimes better than human annotated summary. 
And we have already evaluate consistent and coherent with more fine-grained metrics (Coreference Information, Objective Description, etc).
Thus, we focus on relevance and judge whether the generated summary is informative and relevant.

\paragraph{Overview Scores} Overview score of the summary with the aforementioned metrics.

Table~\ref{tab: submission-result-human-analysis} and Table~\ref{tab: submission-result-human-analysis-tcs} report the scores from the aforementioned metrics. There is not a universal model which performs the best across all of these metrics, instead, each model excels at certain metrics. Overall, TCS\_WITM achieves the best overview score on the public test set, while GoodBai achieves the best overview score on the hidden test set.

\section{Error Analysis}

\begin{figure}[t]
    \centering
    \begin{tabular}{L{7.2cm}}
    \toprule
        {\bf Dialogue:} \\
            \#Person1\#: What time is it, \cgreen{Tom?} \\
            \#Person2\#: Just a minute. It's ten to nine by my watch. \\
            \#Person1\#: Is it? I had no idea it was so late. I must be off now. \\
            \#Person2\#: What's the hurry? \\
            \#Person1\#: I must catch the nine-thirty train. \\
            \cgreen{\#Person2\#: You've plenty of time yet. The railway station is very close. It won't take more than twenty minutes to get there.} \\
        \midrule
        {\bf Gold: }\\
        {\it Summary:} \#Person1\# is catching a train. \cgreen{Tom asks \#Person1\# not to hurry.} \\
        \midrule
        {\bf IITP-CUNI:} \\
        {\it Prediction: }\cred{\#Person1\# and Tom} are in a hurry to catch the nine-thirty train.\\
    \bottomrule
    \end{tabular}
    \caption{An error example where the model fails to distinguish the intent between \cred{\#Person1\# and Tom (\#Person2\#)}.}
    \label{fig:iitp-cuni-example-factual}
\end{figure}

Table~\ref{tab: submission-result-automatic-metrics},~\ref{tab: submission-result-automatic-metrics-tcs},~\ref{tab: submission-result-human-analysis},~\ref{tab: submission-result-human-analysis-tcs} show that the submitted models underperform human beings. Here we analyze some examples where the models make mistakes or fail to capture certain information.

Figure~\ref{fig:iitp-cuni-example-factual} shows an example where the model from IITP-CUNI makes a factual error and fails to reason about who is in the hurry~\footnote{\label{fn:include1summary}We only include one of the gold summaries for demonstration purpose.}. In order to capture the correct relationship, the model needs to reason that ``Tom'' is the name of \#Person2\#, and Tom (\#Person2\#) is asking \#Person1\# not to hurry by saying ``You've plenty of time yet''. However, the model from IITP-CUNI fails in such reasoning processes. This suggests that reasoning about information across the dialogue discourse is challenging.

Figure~\ref{fig:no-cause-example} shows an example where all of the model predictions deviate from the gold summaries. All of the 3 summaries annotated by human beings include the context of \#Person1\# being lost on the way. In contrast, none of the model predictions include this context. This is plausible as the majority of the dialogue is dedicated to \#Person2\# showing \#Person1\# the correct direction, and the model might only capture such salient information in the dialogue. However, the general pattern when human beings summarize is to lay out the cause (context) of an event before telling the event, which is demonstrated in the gold summaries. Thus, there is still a large room for improvement for the model to generate human-like summaries.

Appendix~\ref{appendix: more-error-examples} gives more examples of models making mistakes in terms of the metrics from \S~\ref{sec: human-analysis}.

\section{Conclusion}

We host \emph{DialogSum Challenge} of summarizing daily dialogue conversation at INLG 2021. Our dataset possesses characteristics distinguished from the existing datasets and poses new challenge to the summarization community. There are 4 teams who submit their models during the challenges. An overview of their methods is provided in this report. We evaluate their predictions by automatic metrics and human analysis. Results show that there are still rooms for models to improve the \textsc{Rouge} scores, as well as the quality of the generated summaries in terms of identifying interlocutors' intents, capturing the discourse relation, etc. Besides, we observe the mismatch between \textsc{BERTScore} and human scores, which aligns with the findings by~\citet{hanna2021fine}. Therefore, we advocate to our community to explore automatic metrics that can better align with human scores.

\section*{Acknowledgement}
Yue Zhang is the corresponding author.
We would like to thank Yi Zhang for her help in the human analysis as well as the collection of the hidden test set.
This work receives a support from the Tencent AI Lab Rhino- Bird Focused Research Program.
\bibliography{anthology,custom}
\bibliographystyle{acl_natbib}

\newpage
\appendix

\section{More Error Examples}
\label{appendix: more-error-examples}

Figure~\ref{fig:coref-error-example},~\ref{fig:discourse-relation-error-example},~\ref{fig:objective-error-example},~\ref{fig:intent-error-example},~\ref{fig:low-summarization-and-overview-score} show examples of predicting the wrong coreference information, wrong discourse relation, summarizing with description that is not objective, wrong intent of the interlocutors and with a low standard summarization score as well as a low overview score, respectively.

\begin{figure}
    \centering
    \begin{tabular}{L{7.2cm}}
    \toprule
        {\bf Dialogue:} \\
        \#Person1\#: How long does it take to get to downtown from here? \\
        \#Person2\#: It is 15 minutes ' drive. \\
        \#Person1\#: What companies do we have in our neighborhood? \\
        \#Person2\#: Mitsubishi, HP, IBM and many other famous corporations. \\
        \#Person1\#: Does the 7th floor belong to our company too? \\
        \#Person2\#: Yes, our company has two floors, the 6th and 7th floor in this building. \\
        \#Person1\#: It ' s such a large firm. Do we have our own staff restaurant? \\
        \#Person2\#: Yes, at the end of the hall. \\
        \midrule
        {\bf Gold: }\\
        {\it Summary:} \#Person2\# tells \#Person1\# information about their company and its surroundings.\\
        \midrule
        {\bf UoT:} \\
        {\it Prediction:} \#Person2\# asks \cred{\#Person2\#} about the distance to downtown and the companies in their neighborhood.\\
    \bottomrule
    \end{tabular}
    \caption{An error example for predicting the wrong coreference information.}
    \label{fig:coref-error-example}
\end{figure}

\begin{figure}
    \centering
    \begin{tabular}{L{7.2cm}}
    \toprule
        {\bf Dialogue:} \\
        \#Person1\#: Hi! How are things going with you? \\
        \#Person2\#: I am doing well. What's up with you? \\
        \#Person1\#: Believe it or not, the company I worked for closed down, so I'm out of a job. \\
        \#Person2\#: The same with me. Have you given much thought to what you want to do next? \\
        \#Person1\#: I am not being all that particular right now because I just need to keep a roof over my head. \\
        \#Person2\#: How about interviews? Have you been on any of those yet? \\
        \#Person1\#: I wish that I could get the opportunity to be interviewed. \\
        \#Person2\#: How about the electrician program that they have listed over there? \\
        \#Person1\#: I read about that, and the position sounded great! \\
        \#Person2\#: Let's go see how we can apply for those positions. \\
        \midrule
        {\bf Gold: }\\
        {\it Summary:} \#Person1\# and \#Person2\# are both unemployed. \#Person2\# suggests applying for the electrician program and \#Person1\# agrees. \\
        \midrule
        {\bf IITP-CUNI:} \\
        {\it Prediction:} \#Person1\# tells \#Person2\# \#Person1\# is out of a job \cred{because} \#Person1\# needs to keep a roof over \#Person1\#'s head. \#Person2\# suggests applying for electrician positions. \\
    \bottomrule
    \end{tabular}
    \caption{An error example for predicting the wrong discourse relation.}
    \label{fig:discourse-relation-error-example}
\end{figure}

\begin{figure}
    \centering
    \begin{tabular}{L{7.2cm}}
    \toprule
        {\bf Dialogue:} \\
        \#Person1\#: John dates her seven times a week. \\
        \#Person2\#: Really? That's a straws in the wind. \\
        \#Person1\#: I think so. Maybe he's fallen for her. \\
        \#Person2\#: Yeah. They suit each other. A perfect match between a man and a girl. \\
        \#Person1\#: Right. \\
        \midrule
        {\bf Gold: }\\
        {\it Summary:} \#Person1\# and \#Person2\# think that John and the girl are a perfect match. \\
        \midrule
        {\bf IITP-CUNI:} \\
        {\it Prediction:} \#Person1\# and \#Person2\# talk about John and \cred{the girl he loves}. \\
    \bottomrule
    \end{tabular}
    \caption{An error example for summarization with description that is not objective.}
    \label{fig:objective-error-example}
\end{figure}

\begin{figure}
    \centering
    \begin{tabular}{L{7.2cm}}
    \toprule
        {\bf Dialogue:} \\
        \#Person1\#: Can I help you? \\
        \#Person2\#: I need some stamps for this letter. \\
        \#Person1\#: What kind of stamps do you want? \\
        \#Person2\#: How much do I need for this letter? \\
        \#Person1\#: I must weigh it first. Err... It's five grams over weigh, Do you want to send it as an ordinary or registered letter? \\
        \#Person2\#: I want it registered. How much is it then? \\
        \#Person1\#: Registration plus overnight... err... seven dollars in all. \\
        \#Person2\#: Here's a 10 - dollar bill. \\
        \#Person1\#: Now, your receipt, and the change. \\
        \#Person2\#: Thanks. Good-bye. \\
        \midrule
        {\bf Gold: }\\
        {\it Summary:} \#Person2\# wants to \cgreen{send a letter}. \#Person1\# says it's five grams overweight plus overnight so seven dollars in all. \\
        \midrule
        {\bf GoodBai:} \\
        {\it Prediction:} \#Person1\# helps \#Person2\# \cred{buy some stamps} for a registered letter. \\
    \bottomrule
    \end{tabular}
    \caption{An error example for predicting the wrong intent of the interlocutors.}
    \label{fig:intent-error-example}
\end{figure}

\begin{figure}
    \centering
    \begin{tabular}{L{7.2cm}}
    \toprule
        {\bf Gold: }\\
        {\it Summary:} \#Person1\# and \#Person2\# are talking about the terrorist attacks on 9-11, which was nightmarish for \#Person1\#'s family. \\
        \midrule
        {\bf UoT:} \\
        {\it Prediction:} \#Person1\# and \#Person2\# talk about where they were for the terrorist attacks on 9-11. \#Person1\# was at home with \#Person1\#'s parents in New York City and \#Person1\# didn't see the crash itself \cred{but}\\
    \bottomrule
    \end{tabular}
    \caption{An example with a low standard summarization score and a low overview score. The prediction generated by the model seems unfinished.}
    \label{fig:low-summarization-and-overview-score}
\end{figure}

\end{document}